# Fast and Accurate Point Cloud Registration using Trees of Gaussian Mixtures

Ben Eckart Kihwan Kim Jan Kautz

NVIDIA Research

**Abstract.** Point cloud registration sits at the core of many important and challenging 3D perception problems including autonomous navigation, SLAM, object/scene recognition, and augmented reality. In this paper, we present a new registration algorithm that is able to achieve state-of-the-art speed and accuracy through its use of a hierarchical Gaussian Mixture Model (GMM) representation. Our method constructs a top-down multi-scale representation of point cloud data by recursively running many small-scale data likelihood segmentations in parallel on a GPU. We leverage the resulting representation using a novel PCA-based optimization criterion that adaptively finds the best scale to perform data association between spatial subsets of point cloud data. Compared to previous Iterative Closest Point and GMM-based techniques, our tree-based point association algorithm performs data association in logarithmic-time while dynamically adjusting the level of detail to best match the complexity and spatial distribution characteristics of local scene geometry. In addition, unlike other GMM methods that restrict covariances to be isotropic, our new PCA-based optimization criterion well-approximates the true MLE solution even when fully anisotropic Gaussian covariances are used. Efficient data association, multi-scale adaptability, and a robust MLE approximation produce an algorithm that is up to an order of magnitude both faster and more accurate than current state-of-the-art on a wide variety of 3D datasets captured from LiDAR to structured light.

## 1 Introduction

Point cloud registration is the task of aligning two or more point clouds by estimating the relative transformation between them, and it has been an essential part of many computer vision algorithms such as 3D object matching [1], localization and mapping [2], dense 3D reconstruction of a scene [3], and object pose estimation [4].

Recently point set registration methods [5] have been gaining more importance due to the growing commercial interest of virtual and mixed reality [6], commercial robotics, and autonomous driving applications [7,8]. In most of these applications, massive amounts of 3D point cloud data (PCD) are directly captured from various active sensors (i.e., LiDAR and depth cameras) but at different times under different poses or local coordinate systems. The task of point cloud registration is then to try to find a common coordinate system, which is done by estimating some type of geometric similarity in the point data that can be recovered through optimization over a set of spatial transformations.

One of the oldest and most widely used registration algorithms, Iterative Closest Point (ICP) [9,10], is based on an iterative matching process where point proximity establishes candidate point pair sets. Given a set of point pairs, the rigid transformation that minimizes the sum of squared point pair distances can be calculated efficiently in closed form. ICP and its dozens of variants [11] often fail to produce correct results in many common but challenging scenarios, where the presence of noise, uneven point density, occlusions, or when large pose displacements can cause a large proportion of points to be without valid matches.

Compared to traditional ICP-based approaches, much research has been done on the use of statistical models for registration, which in principle can provide better estimates for outlier rejection, convergence, and geometric matching [12,13,14]. In particular, many statistical methods have been designed around the Expectation Maximization (EM) algorithm [15] as it has been shown that EM generalizes the ICP algorithm under a few basic assumptions [16,17]. Many statistical registration techniques have explicitly utilized this paradigm to deliver better robustness and accuracy [18,19,17,20], but these algorithms tend to be much slower than ICP and often offer only marginal improvement in all but a few specific circumstances. As a result, ICP-based methods are still heavily used in practice for many real-world applications.

Our proposed method falls into the category of GMM-based statistical registration algorithms. We tackle the typical shortcomings of these methods, slow speeds and lack of generality, by adopting an efficient hierarchical construction for the creation of an adaptive multi-scale point matching process. *Efficiency:* The search over multiple scales as a recursive tree-based search produces a highly performant logarithmic-time algorithm that quickly and adaptively finds the most appropriate level of geometric detail with which to match points. *Generality:* By using a data-driven point matching procedure over multiple scales, our proposed algorithm can automatically adapt to many different types of scenes, particularly with real-world data where widely varying sampling sparsity and scene complexity are common. Finally, we introduce a novel Mahalanobis distance approximation resembling ICP's point-to-plane distance minimization metric, which more faithfully approximates the true MLE solution under general anisotropic covariances than previous methods.

## 2 Related Work

Our method builds on previous work in GMM-based methods for registration such as GMM-Reg [31][32], JRMPC [19], and MLMD [28], while also leveraging recent results using hierarchical GMMs for point cloud modeling [33]. By adopting a GMM-based paradigm, we gain robustness in situations of large pose displacement, optimal solutions in the form of maximum likelihood estimates, and an ability to more easily leverage point-level parallelism on GPUs. By aug-menting the GMM into a hierarchy, we can efficiently compress empty space, achieve logarithmic-time matching, and perform robust multi-scale data analysis.

| Method | Mult. Link | Aniso-tropic | Multi-Scale | Data Trans. | Assoc. Complex. | Opt. Complex. |
|---|---|---|---|---|---|---|
| ICP [9] | | | | – | $N^2$ | $N$ |
| SoftAssign [12] | ✓ | | ✓† | – | $N^2$ | $N^2$ |
| EM-ICP [17] | ✓ | | ✓ | kd-tree | $N \log N$ | $N^2$ |
| LM-ICP [21] | | | | grid approx. | $N$ | $N$ |
| KC [22] | ✓ | | | grid approx. | $N$ | $V$ |
| TrICP [23] | | | | voxels | $\frac{N^2}{V}$ | $N$ |
| FICP [24] | | | | kd-tree | $N \log N$ | $N$ |
| G-ICP [16] | | ✓ | | kd-tree | $N \log N$ | $N$ |
| CPD [14] | ✓ | | | FGT | $N$ | $N^2$ |
| ECMPR [20] | ✓ | ✓ | | FGT | $N$ | $N$ |
| GMMReg [25] | ✓ | | ✓† | FGT | $N$ | $N^2$ |
| NDT-P2D [26] | ✓ | ✓ | ✓ | voxels+kd-tree | $N \log V$ | $N$ |
| NDT-D2D [26] | ✓ | ✓ | ✓ | voxels+kd-tree | $V \log V$ | $V$ |
| REM-Seg [27] | ✓ | ✓ | ✓† | GMM | $NJ$ | $N$ |
| MLMD [28] | ✓ | ✓ | | GMM | $NJ$ | $J$ |
| SVR [29] | ✓ | | ✓† | GMM‡ | $N^2 \sim N^3$ | $J$ |
| JRMPC [30] | ✓ | | | GMM | $NJ$ | $J$ |
| **Proposed** | ✓ | ✓ | ✓ | GMM-Tree | $N \log J$ | $\log J \sim J$ |

† Implicitly multi-scale via annealing, ‡ Conversion to GMM via SVM

**Table 1. A Comparison of Registration Methods.** *Multiply Linked*: Many-to-one or many-to-many correspondences, *Anisotropic*: General shape alignment using unrestricted covariance structures, *Multi-Scale*: Registration at multiple levels of granularity, *Data Transform*: Underlying data structure or transform, *Association Complexity*: Complexity of data association problem over all $N$ points (E Step in the case of EM-based methods), *Optimization Complexity*: Size of the optimization problem (M Step in the case of EM-based methods). Assuming both point clouds size $N$, number of voxels/grid points $V$, and number of mixture components $J$.

The earliest statistical methods placed an isotropic covariance around every point in the first set of points and then registered the second set of points to it under an MLE framework (MPM [18], EM-ICP [17], CPD [14,34]). More modern statistical approaches utilize a generative model framework, where a GMM is usually constructed from the points explicitly and registration is solved in an MLE sense using an EM or ECM [35] algorithm (REM-Seg [27], ECMPR [20], JRMPC [19], MLMD [28]), though some utilize a max correlation or $L_2$ distance approach (Kernel Correlation [22], GMM-Reg [31,32], SVR[29], NDT-D2D[36]). Since a statistical framework for point cloud registration tends to be more heavyweight than ICP, techniques such as decimation (EM-ICP [17]), voxelization (NDT methods [36,26]), or Support Vector Machines (SVR [29]) have been used to create smaller or more efficient models, while others have relied on computational tricks such as the Fast Gauss Transform (CPD [14], ECMPR [20]), or have devised ways to exploit point-level parallelism and GPU-computation for increased computational tractability and speed (MLMD [28], parallelized EM-ICP [13]).

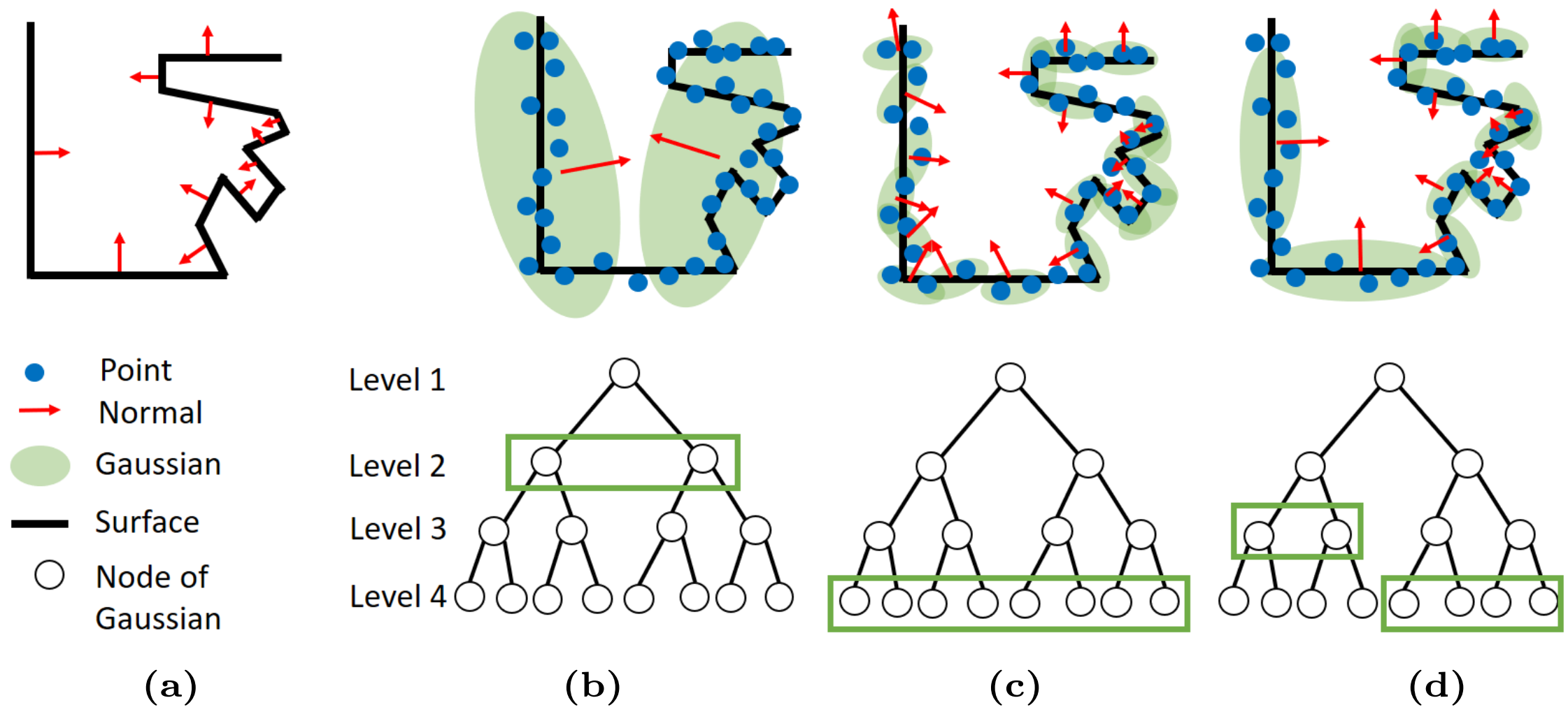

**Fig. 1. Multi-Scale Representation using a Hierarchy of Gaussian Mixtures:** Top-row shows identical geometries (black lines) and associated points (blue circles), which are represented by different levels of Gaussian models (green contour for 1 $\sigma$.) (a) (Top) Ideal Normals (red arrows) on the surfaces, (b) Too coarse (only two Gaussians in Level 2): poor segmentation leads to incorrect normals, which will degrade accuracy when registering points to model, (c) Too fine (using finest level of Gaussian models): over-segmentation leads to erroneous normals as sample noise overtakes real facet geometry (d) Adaptive multi-scale (Mixture of level 3 and level 4 models): point-to-model association can be much more robust when fidelity adaptively changes according to data distribution so that facets can be well-modeled given differing spatial frequencies and sampling densities.

In contrast to these statistical model-based approaches, modern robust variants of point-to-plane ICP (e.g. Trimmed ICP [37], Fractional ICP [24]) are often much faster and sometimes perform nearly as well, especially under real-world conditions [38]. See **Table 1** for a detailed comparison of key registration algorithms utilizing the ICP and GMM paradigms. Our proposed method offers favorable complexity over both classes of algorithms due to its novel use of a GMM-Tree structure, without needing to resort to discretization strategies like the NDT-based methods.

## 3 Registration as Expectation Maximization

The Expectation Maximization (EM) algorithm forms the theoretical foundation for most modern statistical approaches to registration and also generalizes ICP under certain basic assumptions. EM is commonly employed for MLE optimization in the case where directly maximizing the data likelihood for the sought after variable is intractable, but maximizing the expected joint data likelihood conditioned on a set of latent variables is tractable. For the registration case, the sought after variable is the transformation $T$ between point clouds and the latent variables are the point-model associations.

The problem is set up as follows: Given point clouds $\mathcal{Z}_1$ and $\mathcal{Z}_2$, we would like to maximize the data probability of $\mathcal{Z}_2$ under a set of transformations $T$ with respect to a probability model $\boldsymbol{\Theta}_{\mathcal{Z}_1}$ derived from the first point cloud $\mathcal{Z}_1$.

$$\hat{T} = \underset{T}{\operatorname{argmax}}\, p(T(\mathcal{Z}_2)|\hat{\boldsymbol{\Theta}}_{\mathcal{Z}_1}) \tag{1}$$

That is, the most likely estimate of the transformation $\hat{T}$ is the estimate that maximizes the probability that the samples of the transformed point cloud $T(\mathcal{Z}_2)$ came from some probabilistic representation of spatial likelihood (parameterized by $\hat{\boldsymbol{\Theta}}$) derived from the spatial distribution of the first point cloud $\mathcal{Z}_1$. The most common form for parametrizing this probability distribution is through a Gaussian Mixture Model (GMM), whose data probability is defined as a convex combination of $J$ Gaussians weighted by the $J$-component vector $\boldsymbol{\pi}$,

$$p(z|\boldsymbol{\Theta}_{\mathcal{Z}_1}) = \sum_{j=1}^{J} \pi_j \mathcal{N}(z|\boldsymbol{\Theta}_j) \tag{2}$$

The derivation of the probability model $\boldsymbol{\Theta}_{\mathcal{Z}_1}$ may be as simple as statically setting an isotropic covariance around each point in $\mathcal{Z}_1$ (e.g. EM-ICP [17]), or as complicated as framing the search for $\boldsymbol{\Theta}_{\mathcal{Z}_1}$ as a completely separate optimization problem (e.g. SVR [29], MLMD [28]). Regardless of how the model is constructed, however, EM provides an iterative procedure to solve for $T$ through the introduction of a set of latent correspondence variables $\mathcal{C} = \{c_{ij}\}$ that dictate how points $\mathbf{z}_i \in \mathcal{Z}_2$ probabilistically associate to the $J$ subcomponents $\boldsymbol{\Theta}_j$ of the model $\boldsymbol{\Theta}_{\mathcal{Z}_1}$. Intuitively, we can view EM as a statistical generalization of ICP: The E Step estimates data associations, replacing ICP's matching step, while the M Step maximizes the expected likelihood conditioned on these data associations, replacing ICP's distance minimization step over matched pairs.

In the E Step, we use Bayes' rule to calculate expectations over the correspondences. For a particular point $\mathbf{z}_i$, its expected correspondence to $\boldsymbol{\Theta}_j$ ($E[c_{ij}]$) can be calculated as follows,

$$E[c_{ij} = 1] = \frac{\pi_j \mathcal{N}(\mathbf{z}_i|\boldsymbol{\Theta}_j)}{\sum_{k=1}^{J} \pi_k \mathcal{N}(\mathbf{z}_i|\boldsymbol{\Theta}_k)} \tag{3}$$

Generally speaking, larger model sizes (larger $J$) produce more accurate registration results since larger models have more representational fidelity. However, large models produce very slow registration algorithms: Given $N$ points in $\mathcal{Z}_2$, Equation 3 must be calculated $N \times J$ times for each subsequent M Step. For methods that utilize models of size $J \approx O(N)$ (e.g. EM-ICP [17], CPD [14], GMMReg [31]), this causes a data association complexity of $O(N^2)$ and thus these algorithms have problems scaling beyond small point cloud sizes.

To combat this scaling problem, our approach builds from recent advances in fast statistical point cloud modeling via hierarchical generative models by Eckart *et al.* [33]. In this approach, point cloud data is modeled via a GMM-Tree, which is built in a top-down recursive fashion from small-sized Gaussian Mixtures. Their efficient GPU-based approach can produce high-fidelity GMM-Trees in real-time, but given that they were originally designed to optimize reconstructive fidelity and for dynamic occupancy map generation, it is not obvious how to efficiently adapt these models for use in a registration setting. That is, we must derive a way to associate new data to the model and then use the associations to drive an optimization over $T$. As such, we can use their model construction algorithm in order to construct $\Theta_{\mathcal{Z}_1}$ from $Z_1$ (see [33] for details), but we must derive a separate and new EM algorithm to use these GMM-Tree models for registration.

# 4 Hierarchical Gaussian Mixture Mahalanobis Estimation

In this section, we review our proposed approach for hierarchical GMM-based registration under a new EM framework. In Section 4.1 we discuss our new E Step for probabilistic data association that utilizes the GMM-Tree representation for point clouds, and in Section 4.2 we introduce a new optimization criterion to approximate the MLE $T$ for rigid transformations.

## 4.1 E Step: Adaptive Tree Search

Our proposed E Step uses a recursive search procedure to perform probabilistic data association in logarithmic time. We also introduce an early stopping heuristic in order to select the most appropriate scale at which to associate data to the hierarchical model.

The GMM-Tree representation from [33] forms a top-down hierarchy of 8-component GMM nodes, with each individual Gaussian component in a node having its own 8-component GMM child. Thus, a particular node in the GMM-Tree functions in two ways: first, as a probabilistic partition of the data and second, as a statistical description of the data within a partition. We exploit both of these properties in our proposed E Step by using the partitioning information to produce an efficient search algorithm and by using the local data distributions as a scale selection heuristic.

**Logarithmic Search** Each level in the GMM-Tree forms a statistical segmentation at finer levels of granularity and detail. Crucially, the expectation of a point $\mathbf{z}_i$ to a particular Gaussian component $\boldsymbol{\Theta}_j$ is exactly the sum of the expectations of that point to its child GMM. Thus, if we query a parent node's point-model expectation and it falls under a threshold, we can effectively *prune* away all its children's expectations, thus avoiding calculating all $N \times J$ probabilistic associations. Refer to Algorithm 1 for details. In our implementation, we only traverse down the maximum likelihood path at each step. By utilizing the hierarchy in this way, we can recursively search through the tree in logarithmic time ($O(\log J)$) to calculate a point's expectation. This is opposed to previous registration algorithms using traditional GMM's, where a linear search much be performed over all mixture components ($O(J)$) in order to match data to the model.

**Multiscale Adaptivity** Real-world point clouds often exhibit large spatial discrepancies in sampling sparsity and geometric complexity, and so different parts

**Algorithm 1** E Step for Registration

1: **procedure** E_STEP_ADAPTIVE($\mathcal{Z}_2$, $\boldsymbol{\Theta}_{\mathcal{Z}_1}$)<br>
2:     **for** $\mathbf{z}_i \in \mathcal{Z}_2$ **in parallel do**<br>
3:         searchID $\leftarrow -1$, $\gamma \leftarrow \{0,0,0,0,0,0,0,0\}$<br>
4:         **for** $l = 0$ to $L-1$ **do** // L is max tree level<br>
5:             $\mathcal{G} \leftarrow$ Children(searchID) // Children(-1) $\stackrel{\text{def}}{=} \{0..7\}$<br>
6:             **for** $j \in \mathcal{G}$ **do** // for each child in subtree<br>
7:                 $\gamma[j] \propto \pi_j \mathcal{N}(\mathbf{z}_i | \boldsymbol{\Theta}_j)$ // calculate data-model expectation<br>
8:             **end for**<br>
9:             searchID $\leftarrow \text{argmax}_{j \in \mathcal{G}} \gamma[j]$ // Update with most likely association<br>
10:            **if** Complexity($\boldsymbol{\Theta}$[searchID]) $\leq \lambda_c$ **then**<br>
11:                **break** // early stopping heuristic to prune clusters too simple<br>
12:            **end if**<br>
13:        **end for**<br>
14:        // Accumulate $0^{th}$, $1^{st}$, $2^{nd}$ moments $\{M_j^0, M_j^1, M_j^2\}$ for next M_Step<br>
15:        $\{M_j^0, M_j^1, M_j^2\} \leftarrow$Accumulate($M_j^0, M_j^1, M_j^2, \gamma$[searchID], $\mathbf{z}_i$)<br>
16:    **end for**<br>
17:    **return** $\{M_j^0, M_j^1, M_j^2\}$<br>
18: **end procedure**

of the scene may benefit from being represented at different scales when performing point-scene association. Refer to Figure 1 for an overview of this concept. Under a single scale, the point cloud modeling and matching process might succumb to noise or sampling inadequacies if the given modeling fidelity is not appropriate to the local data distribution.

To take advantage of the GMM-Tree multiscale representation and prevent overfitting, we make a check for the current mixture component's geometric complexity and stop early if this condition is not met. This complexity check acts as a heuristic for proper scale selection. We implement our complexity function (Complexity($\cdot$) in Algorithm 1, L10) as $\frac{\lambda_3}{\lambda_1+\lambda_2+\lambda_3}$ for each covariance where $\lambda_1 \geq \lambda_2 \geq \lambda_3$ are its associated eigenvalues. We experimentally set our adaptive threshold, $\lambda_C = 0.01$ for all experiments. This means we terminate the search at a particular scale if the current cluster associated to the point becomes too planar: when 1% or less of its variance occurs along its normal direction. Experimentally, we have found that if we recurse further, we will likely start to chase noise.

Figure 2 shows a graphical depiction of what our adaptive threshold looks like in practice. The Gaussian mixture components break down the point cloud data at a static tree level of 2 ($J = 64$) and 3 ($J = 512$) as compared to an adaptive model that is split into different recursion levels according to a complexity threshold $\lambda_C = 0.01$. The points are color coded according to their expected cluster ownership. Note that the adaptive model has components of both levels of the GMM hierarchy according how smooth or complex the facet geometry is. The ability to adapt to changing levels of complexity allows our M Step to always use a robustly modeled piece of geometry (cf. Figure 1).

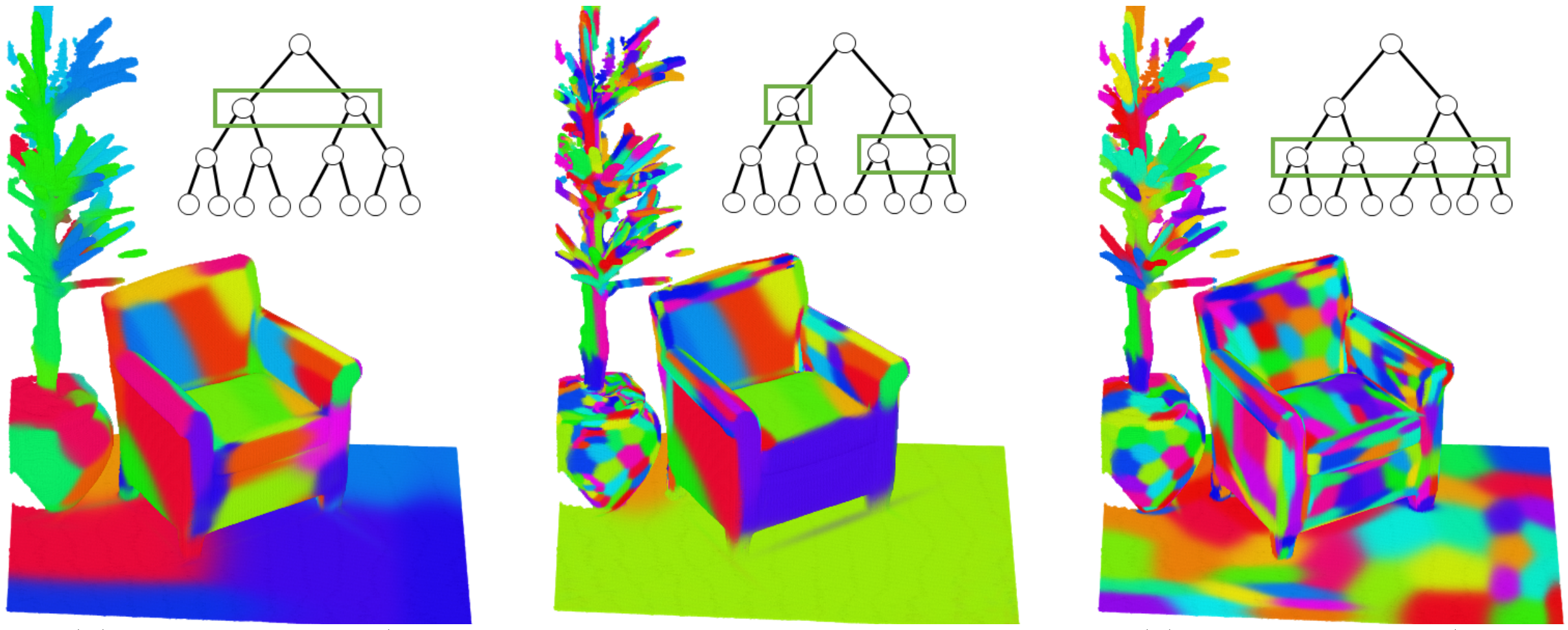

(a) GMM-Tree L2 (64 components) (b) Adaptive L4 ($\lambda_C = 0.01$) (c) GMM-Tree L3 (512 components)

**Fig. 2. Scale Selection using a GMM-Tree** To show qualitatively how scale selection works, we first build a model over a crop (couch, plant, and floor) of the Stanford Scene Lounge dataset [39]. We then associate random colors to each mixture component and color each point according to its data-model expectation. (a) shows this coloring given a static recursion level of 2 in the GMM-Tree, while (c) shows this coloring for a static recursion level of 3. We contrast this with (b), which shows our adaptively scale-selected model containing components at varying levels of recursion depending on the local properties of the mixture components. The scale selection process provides our Mahalanobis estimator (Sec. 4.2) robust component normals, preventing the use of over-fitted or under-fitted mixture components and resulting in a more accurate registration result.

### 4.2 M Step: Mahalanobis Estimation

In this section, we will derive a new M Step for finding the optimal transformation $T$ between a point set $\mathcal{Z}_2$ and an arbitrary GMM $\hat{\boldsymbol{\Theta}}_{\mathcal{Z}_1}$ representing point set $\mathcal{Z}_1$.

First, given $N$ points $\mathbf{z}_i$ and $J$ clusters $\boldsymbol{\Theta}_j \in \hat{\boldsymbol{\Theta}}_{\mathcal{Z}_1}$, we introduce a $N \times J$ set of point-cluster correspondences $\mathcal{C} = \{c_{ij}\}$, so that the full joint probability becomes

$$\ln p(T(\mathcal{Z}), \mathcal{C} | \boldsymbol{\Theta}) = \sum_{i=1}^{N} \sum_{j=1}^{J} c_{ij} \{\ln \pi_j + \ln \mathcal{N}(T(\mathbf{z}_i) | \boldsymbol{\Theta}_j)\} \tag{4}$$

We iterate between E and M Steps. On the E Step, we calculate $\gamma_{ij} \stackrel{\text{def}}{=} E[c_{ij}]$ under the current posterior. On the M Step, we maximize the expected data log likelihood with respect to $T$ while keeping all $\gamma_{ij}$ fixed,

$$\begin{aligned} \hat{T} &= \operatorname*{argmax}_T E_{p(\mathcal{C}|T(\mathcal{Z}),\boldsymbol{\Theta})}[\ln p(T(\mathcal{Z}), \mathcal{C}|\boldsymbol{\Theta})] &(5)\\ &= \operatorname*{argmax}_T \sum_{ij} \gamma_{ij} \{\ln \pi_j + \ln \mathcal{N}(T(\mathbf{z}_i)|\boldsymbol{\Theta}_j)\} &(6)\\ &= \operatorname*{arg\,min}_T \sum_{ij} \gamma_{ij} (T(\mathbf{z}_i) - \boldsymbol{\mu}_j)^T \boldsymbol{\Sigma}_j^{-1} (T(\mathbf{z}_i) - \boldsymbol{\mu}_j) &(7) \end{aligned}$$

Thus, the most likely transformation $T$ between the point sets is the one that minimizes the weighted sum of squared Mahalanobis distances between points of $\mathcal{Z}_2$ and individual clusters of $\boldsymbol{\Theta}_{\mathcal{Z}_1}$, with weights determined by calculating expected correspondences given the current best guess for $\hat{T}$.

As shown mathematically in previous work [20,19,17,28], if we restrict $T$ solely to the set of all rigid transformations ($T = \{R \in SO(3), t_{3\times1}\}$) we can further reduce the double sum over both points and clusters into a single sum over clusters. This leaves us with a simplified MLE optimization criterion,

$$\hat{T} = \arg\min_{T} \sum_{j} \pi_j^* (T(\boldsymbol{\mu}_j^*) - \boldsymbol{\mu}_j)^T \boldsymbol{\Sigma}_j^{-1} (T(\boldsymbol{\mu}_j^*) - \boldsymbol{\mu}_j) \tag{8}$$

where, $\pi_j^* = \frac{\sum_i \gamma_{ij}}{N}$ and $\boldsymbol{\mu}_j^* = \frac{\sum_i \gamma_{ij} \mathbf{z}_i}{\sum_i \gamma_{ij}}$

We can further relate Equation 8 to the weighted moments calculated by the E Step (see Algorithm 1) as follows,

$$\hat{T} = \arg\min_{T} \sum_{j} M_j^0 \left( T\left( \frac{M_j^1}{M_j^0} \right) - \boldsymbol{\mu}_j \right)^T \boldsymbol{\Sigma}_j^{-1} \left( T\left( \frac{M_j^1}{M_j^0} \right) - \boldsymbol{\mu}_j \right) \tag{9}$$

where $M_j^0 = \sum_i \gamma_{ij}$ and $M_j^1 = \sum_i \gamma_{ij} \mathbf{z}_i$.

One can interpret the Mahalanobis distance as a generalization of point-to-point distance where the coordinate system has undergone some affine transformation. In the case of GMM-based registration, each affine transformation is determined by the covariance, or shape, of the cluster to which points are being registered. For example, clusters that are mostly planar in shape (two similar eigenvalues and one near zero) will tend to aggressively pull points toward it along its normal direction while permitting free movement in the plane. This observation should match one's intuition: given that we have chosen a probabilistic model that accurately estimates local geometry, an MLE framework will utilize this information to pull like geometry together as a type of probabilistic shape matching.

By using fully anisotropic covariances, arbitrarily oriented point-to-geometry relations can be modeled. Optimization of Eq. 8 therefore should produce highly accurate transformation estimate. Previous algorithms in the literature, however, have yet to fully leverage this general MLE construction. Simplifications are made either by 1) placing *a priori* restrictions on the complexity of the Gaussian covariance structure (e.g. isotropic only [19] or a single global bandwidth term [17]), or by 2) using approximations to the MLE criterion that remove or degrade this information [28]. The reasons behind both model simplification and MLE approximation are the same: Eq. 8 has no closed form solution. However, we will show how simply reinterpreting the Mahalanobis distance calculation can lead to a highly accurate, novel method for registration.

We first rewrite the inner Mahalanobis distance inside the MLE criterion of Eq. 8 by decomposing each covariance $\boldsymbol{\Sigma}_j$ into its associated eigenvalues $\lambda$ and eigenvectors $\boldsymbol{n}$ using PCA, thereby producing the following equivalence,

$$||T(\boldsymbol{\mu}_j^*) - \boldsymbol{\mu}_j||^2_{\boldsymbol{\Sigma}_j} = \sum_{l=1}^{3} \frac{1}{\lambda_l} (\boldsymbol{n}_l^T (T(\boldsymbol{\mu}_j^*) - \boldsymbol{\mu}_j))^2 \tag{10}$$

Thus, we can reinterpret each cluster's Mahalanobis distance term inside the MLE criterion as a weighted sum of three separate point-to-plane distances. The weights are inversely determined by the eigenvalues, with their associated eigenvectors constituting each plane's normal vector. Going back to the example of a nearly planar Gaussian, its covariance will have two large eigenvalues and one near-zero eigenvalue, with the property that the eigenvectors associated with the larger eigenvalues will lie in the plane and the eigenvector associated with the smallest eigenvalue will point in the direction of its normal vector. Since the weights are inversely related to the eigenvalues, we can easily see that the MLE criterion will mostly disregard any point-to-$\boldsymbol{\mu}_j$ distance inside its plane (that is, along the two dominant PCA axes) and instead disproportionately focus on minimizing out-of-plane distances by pulling nearby points along the normal to the plane.

We can see that by plugging in this equivalence back into Eq. 8, we arrive at the following MLE criterion,

$$\hat{T} = \arg\min_{T} \sum_{j=1}^{J} \sum_{l=1}^{3} \frac{\pi_j^*}{\lambda_{j_l}} (\boldsymbol{n}_{j_l}^T (T(\boldsymbol{\mu}_j^*) - \boldsymbol{\mu}_j))^2 \tag{11}$$

where the set of $\boldsymbol{n}_{j_l}, l = 1..3$ represent the 3 eigenvectors for the $j$th Gaussian (anisotropic) covariance, and $\lambda_{j_l}$ the associated eigenvalues.

We have transformed the optimization from the minimization of a weighted sum of $J$ squared Mahalanobis distances to an equivalent minimization of a weighted sum of $3J$ squared point-to-plane distances. In doing so, we arrive at a form that can be leveraged by any number of minimization techniques previously developed for point-to-plane ICP [10]. Note that unlike traditional point-to-plane methods, which usually involve the computationally difficult task of finding planar approximations over local neighborhoods at every point and sometimes also for multiple scales [40,41], the normals in Eq. 12 are found through a very small number of 3x3 eigendecompositions (typically $J \leq 1000$ for even complex geometric models) over the model covariances, with appropriate scales chosen through our proposed recursive search over the covariances in the GMM-Tree (Sec 4.1).

We solve Equation 12 using the linear least squares technique described by Low for point-to-plane ICP optimization [42], which we adapt into a weighted form (a derivation is provided in our supplementary material). The only approximation required is a linearization of $R$ using the small-angle assumption. In practice, this is a fair assumption to use since GMM-based registration methods are local and thus diverge for large pose displacements anyway.

## 5 Speed vs Accuracy

For every registration algorithm, there is an inherent trade-off between accuracy and speed. To explore how different registration algorithms perform under various accuracy/speed trade-offs, we have designed a synthetic experiment using the Stanford Bunny. We take 100 random 6DoF transformations of the bunny and then run each algorithm over the same group of random point subsets of increasing cardinality. Our method of obtaining a random transformation is to sample each axis of rotation uniformly from [-15,15] degrees and each translation uniformly from [-0.05, 0.05] (roughly half the extent of the bunny). We can then plot speed vs accuracy as a scatter plot in order to see how changing the point cloud size (a proxy for model complexity) affects the speed vs accuracy tradeoff.

The algorithms and code used in the following experiments were either provided directly by the authors (JRMPC, ECMPR, NDT-D2D, NDT-P2D, SVR, GMMReg), taken from popular open source libraries (libpointmatcher for TrICP-pt2pt, TrICP-pt2pl, FICP), or are open source re-implementations of the original algorithms with various performance optimizations (EM-ICP-GPU, SoftAssign-GPU, ICP-OpenMP, CPD-C++). Links to the sources can be found in the supplementary material. Parameters were set for all algorithms according to what was recommended by the authors and/or by the software. All our experiments were run on Intel Core i7-5920K and NVIDIA Titan X.

In order to test how each design decision affects the performance of the proposed algorithm, we test against three variants:

1. **Adaptive L**$n$: The full algorithm proposed in this paper. Adaptive multi-scale data association using a GMM-Tree that was constructed up to a max recursion level of $n$.
2. **GMM-Tree L**$n$: Here we use the same GMM-Tree representation for logarithmic time data association, but without multi-scale adaptivity ($\lambda_c = 0$). The tree is constructed up to a max recursion level of $n$. By comparing *GMM-Tree* to *Adaptive*, we can see the benefits of stopping our recursive search according to data complexity.
3. **GMM J=**$n$: This variant forgoes a GMM-Tree representation and uses a simple, fixed complexity, single-level GMM with $n$ mixture components. It is therefore similar to other fixed complexity GMM-based registration approaches to data representation (e.g. [17,31,28,19]). Thus, both recursive (logarithmic) data-association and adaptive complexity cannot be used. However, this variant is still well-optimized for the GPU and can still use our new PCA-based MLE optimization. Comparing this approach (*GMM*) to the tree-based representations (*GMM-Tree* and *Adaptive*) shows how much the tree-based data representation affects registration performance over just using our new MLE optimization technique.

Figure 3(a) shows each algorithm's speed vs accuracy trade-off by plotting registration error vs time elapsed. The lower left corner is best (both fast and accurate). One can quickly see how different classes of algorithms clearly dominate each other on the speed/accuracy continuum. For additional clarity, Figure

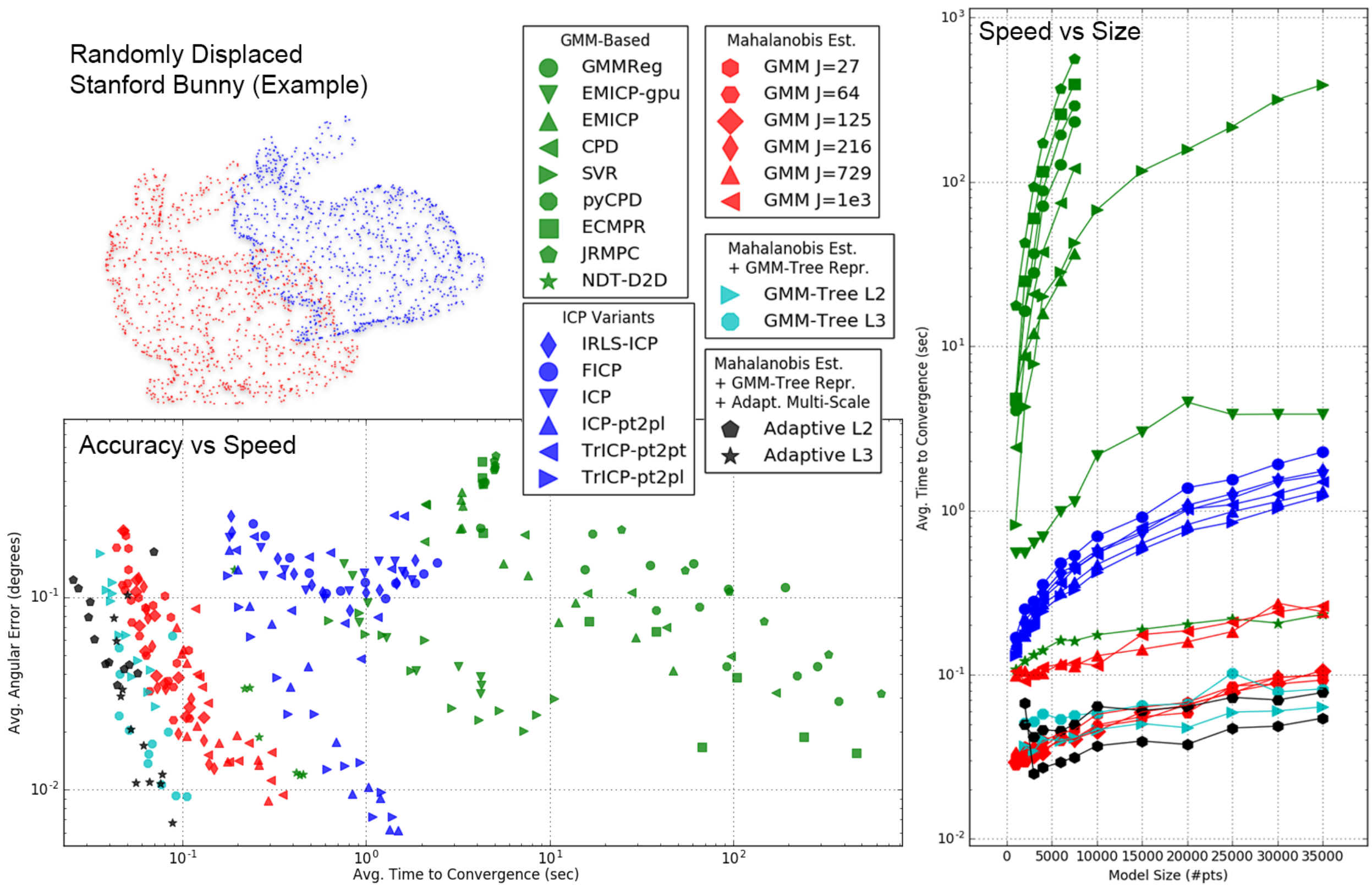

**(a) Accuracy vs Speed** **(b) Speed vs Size**

**Fig. 3.** Each data point represents a particular algorithm's average speed and accuracy when registering together randomly transformed Stanford Bunnies. We produce multiple points for each algorithm at different speed/accuracy levels by applying the methods multiple times to different sized point clouds. The lower left corner shows the fastest and most accurate algorithms for a particular model size. Our proposed algorithms (black, cyan, and red) tend to dominate the bottom left corner, though robust point-to-plane ICP methods sometimes produce more accurate results, albeit at much slower speeds (e.g. Trimmed ICP).

3(b) explicitly plots the time scaling of each registration method as a function of point cloud size. For both timing and accuracy, one can see that, roughly speaking, our adaptive tree formulation performs the best, followed by our non-adaptive tree formulation, followed by our non-adaptive non-tree formulation, then ICP-based variants, and then finally previous GMM-based variants (black > cyan > red > blue > green).

It should be noted that even though our proposed algorithms (black, cyan, and red) tend to dominate the lower left corner of Figure 3(a), certain robust point-to-plane ICP methods sometimes produce more accurate results, albeit at much slower speeds. See for example in Figure 3 that some point-to-plane ICP results were less than $10^{-2\circ}$ angular error and near 1 second convergence time. We estimate that this timing gap might be decreased given a good GPU-optimized robust planar ICP implementation, though it is unclear if the neighborhood-based planar approximation scheme used by these algorithms could benefit from GPU parallelization as much as our proposed Expectation Maximization approach, which is designed to be almost completely data parallel at the point

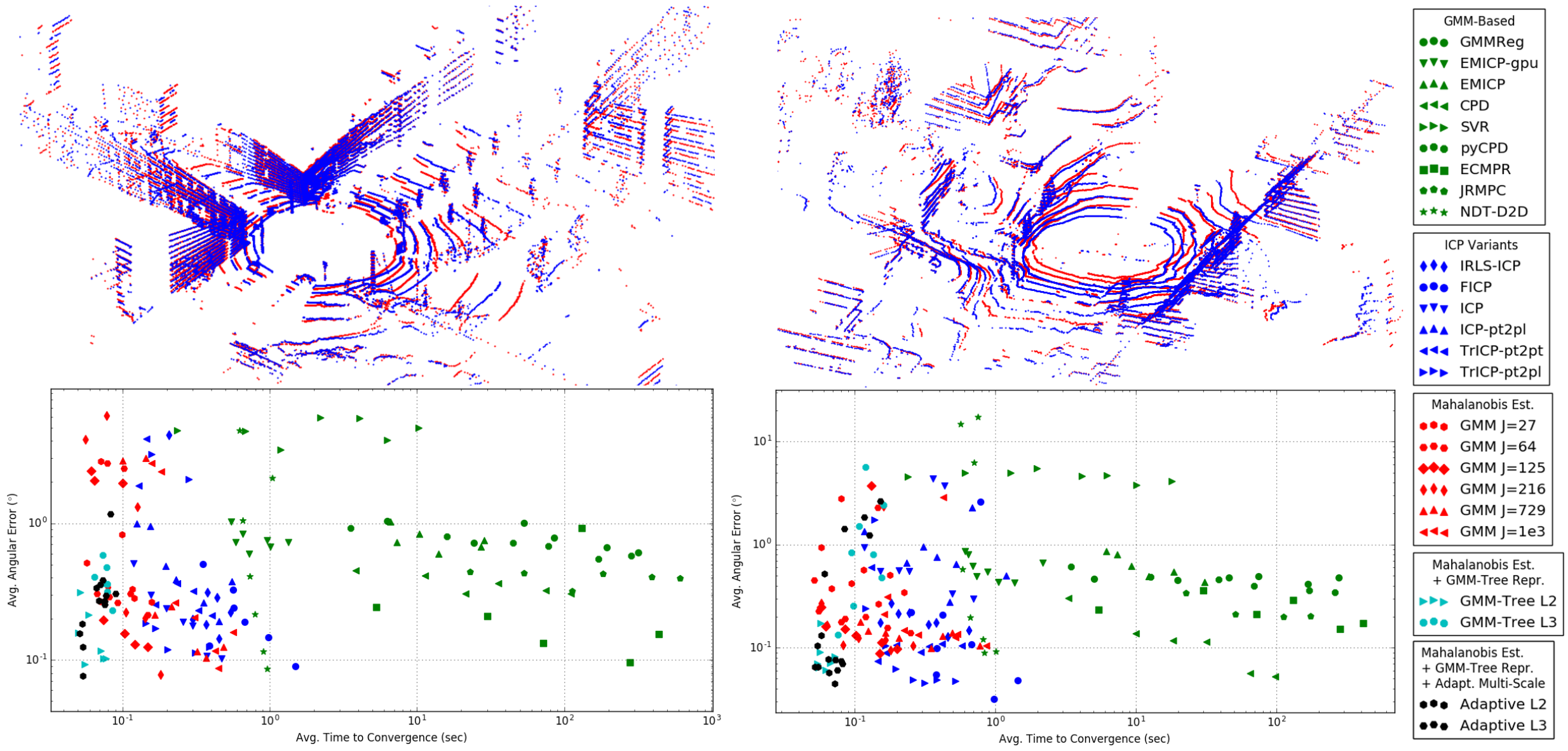

**(a)** Urban scene with many rectilinear structures

**(b)** Snowy, hilly terrain with few features level. However, if computation time is not a constraint for a given application (e.g. offline approaches), we would recommend trying both types of algorithms (our model-based approach vs a robust planar ICP-based approach) to see which provides the best accuracy.

**Fig. 4.** Speed vs accuracy tests for two types of real-world LiDAR frames with very different sampling properties from the Stanford Bunny. In general, similar results are obtained as in Figure 3.

For completeness, we repeated the test with two frames of real-world Lidar data, randomly transformed and varyingly subsampled as before in order to obtain our set of speed/accuracy pairs. The results are shown in Figure 4. As in Fig. 3(a), the bottom left corner is most desirable (both fast and accurate), our methods shown in red, teal, and black. Given that the bunny and LiDAR scans have very different sampling properties, a similar outcome for all three tests shows that the relative performance of the proposed approach isn't dependent on evenly sampled point clouds.

## 6 Evaluation on Real-World Data

**Lounge Dataset** In this test, we calculate the frame-to-frame accuracy on the Stanford Lounge dataset, which consists of range data produced by moving a handheld Kinect around an indoor environment [39]. We register together every 5th frame for the first 400 frames, each downsampled to 5000 points. To measure the resulting error, we calculate the average Euler angle deviation from ground truth. Refer to Table 2(a) for error and timing. All our experiments were run on Intel Core i7-5920K and NVIDIA Titan X. We chose to focus on rotation error since this was where the largest discrepancies were found among algorithms. The best performing algorithm we tested against, Trimmed ICP with point-to-plane distance error minimization, had an average Euler angle error of 0.54 degrees and

| Method | Ang. Error (°) | Speed (fps) |
|---|---|---|
| CPD | 2.11 | 0.18 |
| GMMReg | 3.02 | .04 |
| NDT-D2D | 14.25 | 11.76 |
| FICP | 1.44 | 4.9 |
| ICP | 7.29 | 2.58 |
| IRLS-ICP | 2.29 | 7.1 |
| EMICP | 10.47 | 1.44 |
| SVR | 2.67 | 0.35 |
| ECMPR | 2.21 | 0.059 |
| JRMPC | 8.27 | 0.042 |
| TrICP-pt2pl | 0.54 | 8.4 |
| TrICP-pt2pt | 1.26 | 5.5 |
| ICP-pt2pl | 2.24 | 7.6 |
| **GMM-Tree L2** | **0.77** | **31.3** |
| **GMM-Tree L3** | **0.48** | **20.4** |
| **GMM-Tree L4** | **0.56** | **14.2** |
| **Adaptive L2** | **0.76** | **29.6** |
| **Adaptive L3** | **0.46** | **19.8** |
| **Adaptive L4** | **0.37** | **14.5** |

**(a) Lounge Dataset**

| Method | Ang. Error (°) | Trans. Error (cm) | Speed (fps) |
|---|---|---|---|
| CPD | 0.15 | 17.2 | 0.004 |
| GMMReg | 0.73 | 102.1 | 0.22 |
| NDT-D2D | 0.17 | 16.0 | 0.88 |
| FICP | 0.15 | 35.1 | 1.01 |
| ICP | 0.26 | 15.0 | 1.35 |
| IRLS-ICP | 0.15 | 14.7 | 1.28 |
| EMICP | 0.99 | 103.1 | 2.05 |
| SVR | 0.21 | 39.1 | 0.27 |
| ECMPR | 0.31 | 24.1 | 0.21 |
| JRMPC | 0.60 | 73.1 | 0.05 |
| TrICP-pt2pl | 0.15 | 43.2 | 1.74 |
| TrICP-pt2pt | 0.21 | 66.2 | 1.75 |
| ICP-pt2pl | 0.27 | 7.5 | 1.48 |
| **GMM-Tree L2** | **0.11** | **12.5** | **39.34** |
| **GMM-Tree L3** | **0.18** | **23.9** | **21.41** |
| **GMM-Tree L4** | **0.20** | **29.5** | **15.00** |
| **Adaptive L2** | **0.12** | **10.0** | **39.20** |
| **Adaptive L3** | **0.15** | **8.8** | **22.82** |
| **Adaptive L4** | **0.15** | **9.2** | **16.91** |

**(b) LiDAR Dataset**

**Table 2. Comparison of Registration Methods for the Lounge and LiDAR Datasets** Timing results for both datasets include the time to build the GMM-Tree. Errors are frame-to-frame averages. Speed given is in average frames per second that the data could be processed (note that the sensor outputs data frames at 30 Hz for the Lounge data and roughly 10 Hz for the LiDAR data).

took on average 119 ms to converge. Our best algorithm, the adaptive algorithm to a max depth of 3, had an average Euler angle error of 0.46 degrees and took on average less than half the time (50.5 ms) to converge. Also, note that our times include the time to build the model (the GMM-Tree), which could have other benefits for applications that need to utilize such a model for tasks like loop closure detection, or map building. The accuracy of our proposed methods on this dataset is comparable with the best ICP variants, but at roughly twice the speed.

See Figure 7 for the error distribution over these frames as a set of histograms. We show the error histogram as a few misaligned results can potentially skew the average error, while the histogram gives a more complete picture of both the accuracy and robustness of the method. Note that the previous GMM-based methods (in green) are often prohibitively slow for most applications, while ICP variants perform better. The perform of our proposed methods on this dataset is near to the accuracy of the best ICP variants, but at roughly twice the speed.

**Velodyne LiDAR Dataset** We performed frame-to-frame registration on an outdoor LiDAR dataset using a Velodyne (VLP-16) LiDAR and overlaid the results in a common global frame. See Figure 5 for a qualitative depiction of the result. Table 2(b) summarizes the quantitative results from Figure 5 in an easier to read table format.

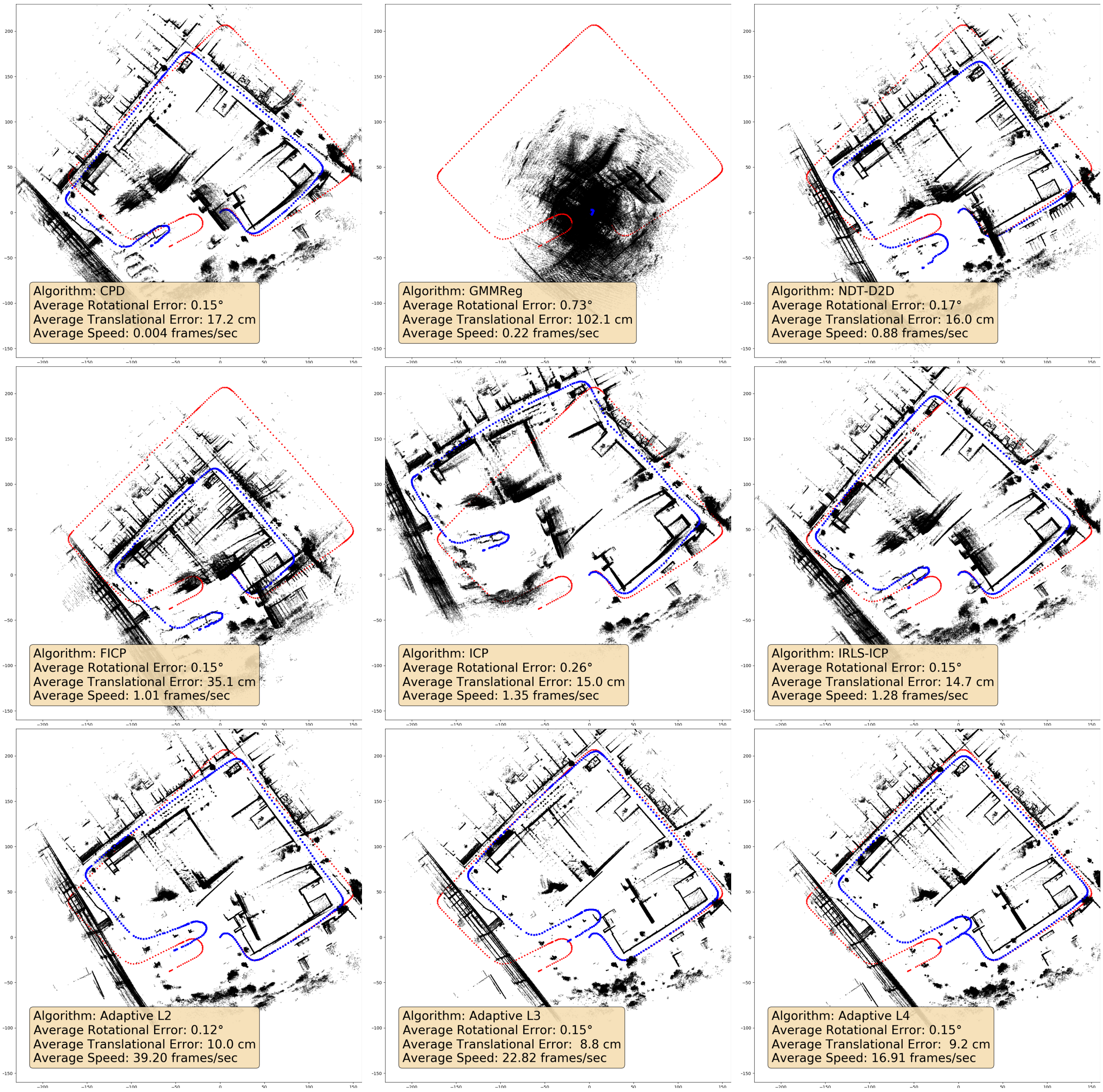

**Fig. 5. Frame-to-Frame Registration with Outdoor LiDAR Dataset:** Ground truth path shown in red, calculated path shown in blue. Each frame of LiDAR data represents a single sweep. We register successive frames together and concatenate the transformation matrices in order to plot the results in a single coordinate system. Note that drift is expected over such long distances as we perform no loop closures. The top row are related GMM-Based methods, the middle row shop modern ICP implementations, and the bottom row show three of our proposed adaptive GMM-Tree methods at three different max recursion levels. For our methods, the timing results include the time to build the recursive GMM model. GMM-Based methods generally perform slowly and GMMReg in particular diverged. ICP-based methods fared better in our testing, though our proposed methods show an order of magnitude improvement in speed while beating or competing with other state-of-the-art in accuracy.

In Figure 5, the ground truth path is shown in red, and the calculated path is shown in blue. Since there is no loop closures, the error is expected to compound and cause drift over time. However, despite the compounding error, the bottom row of Figure 5 (and correspondingly, the bottom three line items of Table 2(b)) shows that the proposed methods can be used for fairly long distances (city blocks), without the need for any odometry (e.g. INS or GPS) or loop closures.

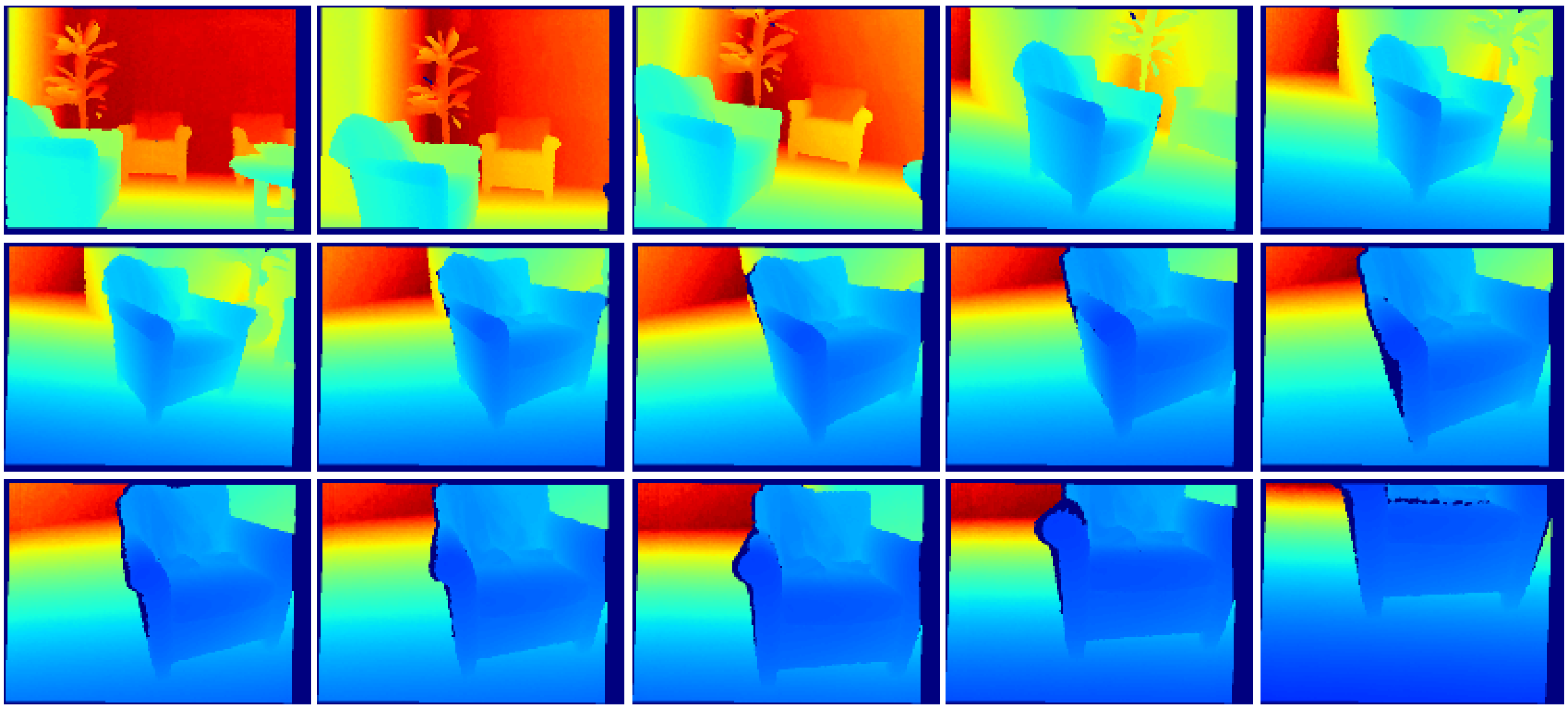

**Fig. 6.** Example of lounge depth stream from which we produce input point clouds for our frame-to-frame registration comparison.

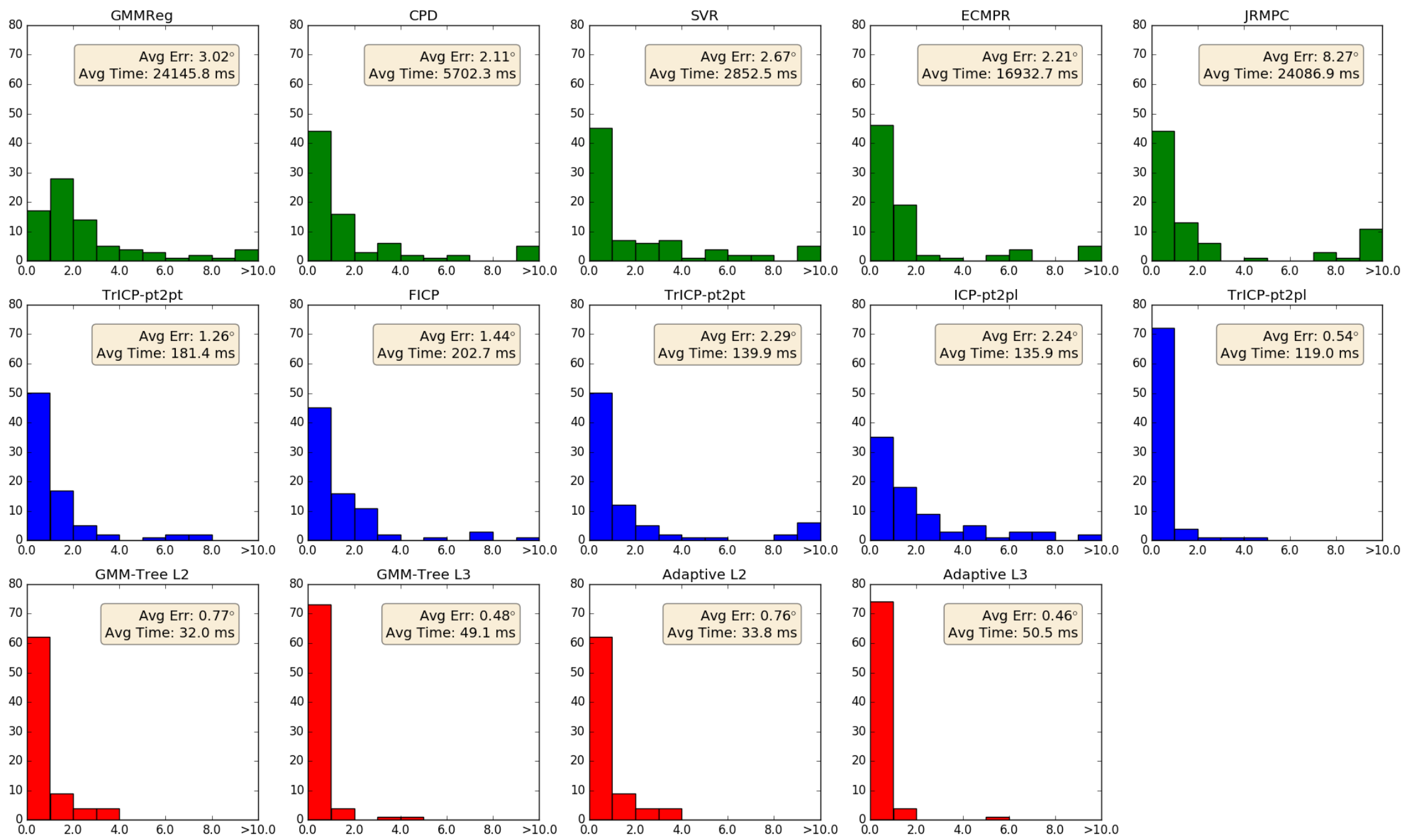

**Fig. 7. Lounge Error Distributions** Histograms of frame-to-frame error as measured by average Euler angular deviation from ground truth, for the Lounge dataset. Each row shows a different type of method: the top row are previous GMM-based methods, the middle row are robust ICP variants, and the bottom row are our proposed algorithms. Note that while some ICP variants achieve good accuracy, our methods are roughly twice as fast. This stands in contrast to previous GMM-based methods, which are prohibitively slow.

This particular LiDAR sensor (Velodyne VLP-16) outputs point cloud sweeps at roughly 10 Hz with an average point cloud size of 13,878 points. Thus, our methods achieve faster than real-time speeds (17-39 Hz), while the state-of-the-art ICP methods are an order of magnitude slower ($\approx 1$ fps). Also, note that our times include the time to build the model (the GMM-Tree), which could be utilized for other concurrent applications besides registration (e.g. probabilistic mesh visualization or loop detection).

To reiterate from the description given in the main paper, the results from the LiDAR test depict single frame-to-frame registration results, where each new result's calculated transformation is concatenated with all the previous so as to put all data into a single global frame. Since there is no loop closures, the error is expected to be compounding, eventually causing extreme drift over time. However, even despite the compounding error, the bottom row of Figure 5 shows that the proposed methods can be used for fairly long distances (city blocks), without the need for any odometry (e.g. INS or GPS) or loop closures.

## 7 Conclusion

To conclude, we proposed a new model-based point cloud registration algorithm that uses a tree of Gaussian mixtures as our data representation. Our proposed technique uses this modeling hierarchy in order to adaptively and efficiently perform association between the model and point data. Framing point association as a recursive tree search results in orders of magnitude speed-up relative to traditional GMM-based approaches that must perform these associations linearly. To further take advantage of the chosen multi-scale anisotropic GMM representation, we introduce a new approximation scheme using PCA that reduces the MLE optimization to a weighted point-to-plane measure. We tested our proposed methods along with a large variety of state-of-the-art registration methods and found that our approach is appropriate for many types of point cloud data and often an order of magnitude faster than current state-of-the-art while producing similar or greater accuracy. The implementation of our method can be found at `http://placeholder`.

# Appendices

## Related Work: Sources

In this section, we supply all the sources used for the related work we tested against in the main paper. We used the author's suggested parameters for their software unless otherwise noted.

**SVR** https://sites.google.com/view/djcampbell/research-software
*Note:* Annealing disabled (resulted in better speeds without much loss in accuracy)

**GPU-Accelerated SoftAssign, EM-ICP and ICP** https://github.com/tttamaki/cuda_emicp_softassign

**ECMPR** https://team.inria.fr/perception/research/ecmpr/

**CPD** (Python): https://github.com/siavashk/pycpd
(C++): https://github.com/gadomski/cpd

**JRMPC** https://team.inria.fr/perception/research/jrmpc/
*Note:* JRMPC is designed for multi-frame batch registration, though we tested it as a frame-to-frame registration method in the paper. We decided to include it despite this due to its similarity to our method.

**Trimmed Point-to-Plane ICP, Trimmed Point-to-Point ICP, Fractional ICP, ICP with Iteratively Reweighted Least Squares, Point-to-Point ICP** https://github.com/ethz-asl/libpointmatcher

*Note:* These algorithms were each implemented using the *libpointmatcher* library. In every case, *IdentityDataPointsFilter* was used as the sole data points filter. That is, no subsampling was performed for any of these methods. We based our point-to-point and point-to-plane configurations off of the sample yaml files found in the *evaluations/official_solutions* directory (Besl92_pt2point.yaml and Chen91_pt2plane.yaml).

1. Trimmed Point-to-Plane ICP: Trimmed distance ratio = 0.7, 7-NN for normal calculation
2. Trimmed Point-to-Point ICP: Trimmed distance ratio = 0.75
3. Point-to-Plane ICP: 10-NN for normal calculation
4. Fractional ICP: Trimmed distance filter minRatio = 0.05
5. IRLS-ICP: Robust Welsch outlier filter scale = 5.0

**GMMReg** https://github.com/bing-jian/gmmreg

**NDT-D2D, NDT-P2D** http://wiki.ros.org/ndt_registration

*Note:* We utilized slightly larger voxel sizes on the LiDAR dataset than was default so as to avoid out of memory errors.

1. NDT-P2D: 4-level multiscale (voxel sizes of 0.5, 1, 2, and 4) (to avoid out of memory errors on LiDAR dataset)
2. NDT-D2D: 4-level multiscale (voxel sizes of 0.5, 1, 2, and 4) (to avoid out of memory errors on LiDAR dataset)

## Expanded Derivation: Mahalanobis Estimation (Sec 4.2)

In this final section, we derive the full weighted least squares solution for the Mahalanobis approximator given in Section 4.2.

Recall that with a GMM of size $J$, the MLE optimization is as follows,

$$\hat{T} = \arg\min_{T} \sum_{j=1}^{J} \sum_{l=1}^{3} \frac{\pi_j^*}{\lambda_{j_l}} (\boldsymbol{n}_{j_l}^T (T(\boldsymbol{\mu}_j^*) - \boldsymbol{\mu}_j))^2 \tag{12}$$

where the set of $\boldsymbol{n}_{j_l}, l = 1..3$ represent the 3 eigenvectors for the $j$th Gaussian (anisotropic) covariance, and $\lambda_{j_l}$ the associated eigenvalues.

To set up the weighted linear least squares solution, we will first define three vectors of weights,

$$\mathbf{w_1} = \begin{bmatrix} \sqrt{\frac{\pi_1^*}{\lambda_{1_1}}} \\ \sqrt{\frac{\pi_2^*}{\lambda_{2_1}}} \\ \sqrt{\frac{\pi_3^*}{\lambda_{3_1}}} \\ \vdots \\ \sqrt{\frac{\pi_J^*}{\lambda_{J_1}}} \end{bmatrix}, \mathbf{w_2} = \begin{bmatrix} \sqrt{\frac{\pi_1^*}{\lambda_{1_2}}} \\ \sqrt{\frac{\pi_2^*}{\lambda_{2_2}}} \\ \sqrt{\frac{\pi_3^*}{\lambda_{3_2}}} \\ \vdots \\ \sqrt{\frac{\pi_J^*}{\lambda_{J_2}}} \end{bmatrix}, \mathbf{w_3} = \begin{bmatrix} \sqrt{\frac{\pi_1^*}{\lambda_{1_3}}} \\ \sqrt{\frac{\pi_2^*}{\lambda_{2_3}}} \\ \sqrt{\frac{\pi_3^*}{\lambda_{3_3}}} \\ \vdots \\ \sqrt{\frac{\pi_J^*}{\lambda_{J_3}}} \end{bmatrix} \tag{13}$$

Next we define the following three vectors,

$$\mathbf{b_1} = \begin{bmatrix} \boldsymbol{n}_{1_1}^T(\boldsymbol{\mu}_1^* - \boldsymbol{\mu}_1) \\ \boldsymbol{n}_{2_1}^T(\boldsymbol{\mu}_2^* - \boldsymbol{\mu}_2) \\ \boldsymbol{n}_{3_1}^T(\boldsymbol{\mu}_3^* - \boldsymbol{\mu}_3) \\ \vdots \\ \boldsymbol{n}_{J_1}^T(\boldsymbol{\mu}_J^* - \boldsymbol{\mu}_J) \end{bmatrix}, \mathbf{b_2} = \begin{bmatrix} \boldsymbol{n}_{1_2}^T(\boldsymbol{\mu}_1^* - \boldsymbol{\mu}_1) \\ \boldsymbol{n}_{2_2}^T(\boldsymbol{\mu}_2^* - \boldsymbol{\mu}_2) \\ \boldsymbol{n}_{3_2}^T(\boldsymbol{\mu}_3^* - \boldsymbol{\mu}_3) \\ \vdots \\ \boldsymbol{n}_{J_2}^T(\boldsymbol{\mu}_J^* - \boldsymbol{\mu}_J) \end{bmatrix}, \mathbf{b_3} = \begin{bmatrix} \boldsymbol{n}_{1_3}^T(\boldsymbol{\mu}_1^* - \boldsymbol{\mu}_1) \\ \boldsymbol{n}_{2_3}^T(\boldsymbol{\mu}_2^* - \boldsymbol{\mu}_2) \\ \boldsymbol{n}_{3_3}^T(\boldsymbol{\mu}_3^* - \boldsymbol{\mu}_3) \\ \vdots \\ \boldsymbol{n}_{J_3}^T(\boldsymbol{\mu}_J^* - \boldsymbol{\mu}_J) \end{bmatrix} \tag{14}$$

We denote the set of $J$ eigenvectors associated with each covariance's $l$th eigenvalue as $\boldsymbol{n}_{\lambda_l}$. Note that $\odot$ is the elementwise product. We weight the normals and vertically stack $\boldsymbol{\mu}_j^*$ as follows,

$$\mathbf{n} = \begin{bmatrix} [\mathbf{w_1}\ \mathbf{w_1}\ \mathbf{w_1}] \odot \boldsymbol{n}_{\lambda_1} \\ [\mathbf{w_2}\ \mathbf{w_2}\ \mathbf{w_2}] \odot \boldsymbol{n}_{\lambda_2} \\ [\mathbf{w_3}\ \mathbf{w_3}\ \mathbf{w_3}] \odot \boldsymbol{n}_{\lambda_3} \end{bmatrix}, \hat{\boldsymbol{\mu}}^* = \begin{bmatrix} \boldsymbol{\mu}_j^* \\ \boldsymbol{\mu}_j^* \\ \boldsymbol{\mu}_j^* \end{bmatrix} \tag{15}$$

If we use the small angle assumption to linearize rotation, we are left with the following overconstrained linear system, where $\{\alpha, \beta, \gamma\}$ are the rotation angles and $\{t_x, t_y, t_z\}$ are the optimal translation parameters.

$$[\hat{\boldsymbol{\mu}}^* \times \mathbf{n}\ \mathbf{n}] \begin{bmatrix} \alpha \\ \beta \\ \gamma \\ t_x \\ t_y \\ t_z \end{bmatrix} = \begin{bmatrix} \mathbf{w_1} \odot \mathbf{b_1} \\ \mathbf{w_2} \odot \mathbf{b_2} \\ \mathbf{w_3} \odot \mathbf{b_3} \end{bmatrix} \tag{16}$$

Equation 16 can be solved with any standard linear least squares algorithm.